\documentclass[11pt,a4paper]{article}

\usepackage[utf8]{inputenc}
\usepackage[T1]{fontenc}
\usepackage{lmodern}
\usepackage[margin=1in]{geometry}
\usepackage{graphicx}
\graphicspath{{../results/figures/}{./}{figures/}}
\usepackage{booktabs}
\usepackage{amsmath}
\usepackage{url}
\usepackage[hidelinks]{hyperref}
\usepackage{xcolor}

\newcommand{\BaselineP}{0.318}
\newcommand{\Chance}{0.25}
\newcommand{\ConsEnLlama}{0.670}
\newcommand{\ConsEnMedGemma}{0.578}
\newcommand{\ConsEnQwen}{0.676}

\newcommand{\ConsHaLlama}{0.210}
\newcommand{\ConsHaMedGemma}{0.408}
\newcommand{\ConsHaQwen}{0.084}
\newcommand{\EFormat}{0.04}

\newcommand{\ELocChosen}{10}
\newcommand{\ELocGold}{6}
\newcommand{\ELocStem}{18}
\newcommand{\ENRows}{50}
\newcommand{\EReason}{0.28}

\newcommand{\ETerm}{0.58}
\newcommand{\ETermHi}{0.71}
\newcommand{\ETermLo}{0.44}
\newcommand{\ETermN}{29}
\newcommand{\ETextTotal}{34}
\newcommand{\ETrans}{0.10}

\newcommand{\EUntranslated}{22}
\newcommand{\EnLlama}{0.618}
\newcommand{\EnMedGemma}{0.550}
\newcommand{\EnQwen}{0.526}
\newcommand{\ExclDental}{1318}
\newcommand{\ExclMultiAnswer}{894}
\newcommand{\FailAllPromptEcho}{52.9}
\newcommand{\FailAllPromptEchoN}{127}
\newcommand{\FailAllRepetition}{0.4}
\newcommand{\FailAllRepetitionN}{1}
\newcommand{\FailAllSwahili}{32.5}
\newcommand{\FailAllSwahiliN}{78}
\newcommand{\FailAllTruncated}{5.0}
\newcommand{\FailAllTruncatedN}{12}
\newcommand{\FailAnyPromptEcho}{43.8}
\newcommand{\FailAnyPromptEchoN}{63}
\newcommand{\FailAnyRepetition}{0.0}
\newcommand{\FailAnyRepetitionN}{0}
\newcommand{\FailAnySwahili}{30.6}
\newcommand{\FailAnySwahiliN}{44}
\newcommand{\FailAnyTruncated}{4.2}
\newcommand{\FailAnyTruncatedN}{6}
\newcommand{\FailAtBudget}{12}
\newcommand{\FailChains}{240}
\newcommand{\FailMedianTokens}{87}
\newcommand{\FailParsedCodeSwitch}{35.4}

\newcommand{\FailParsedPromptEcho}{66.7}

\newcommand{\FailTokenBudget}{384}

\newcommand{\FailUnparsed}{144}
\newcommand{\FallbackItems}{95}
\newcommand{\GapHiLlama}{39.4}
\newcommand{\GapHiMedGemma}{24.4}
\newcommand{\GapHiQwen}{45.4}
\newcommand{\GapLlama}{34.2}
\newcommand{\GapLoLlama}{28.8}
\newcommand{\GapLoMedGemma}{14.4}
\newcommand{\GapLoQwen}{35.2}
\newcommand{\GapMedGemma}{19.4}
\newcommand{\GapQwen}{40.4}
\newcommand{\GateAcc}{0.079}
\newcommand{\GateAccParsed}{0.198}
\newcommand{\GateAccParsedHi}{0.289}
\newcommand{\GateAccParsedLo}{0.131}
\newcommand{\GateItems}{60}
\newcommand{\GateK}{4}

\newcommand{\GatePairs}{17}
\newcommand{\GateParse}{0.400}
\newcommand{\GateParsedN}{96}
\newcommand{\GateProjected}{122}
\newcommand{\GateSamples}{240}
\newcommand{\GateThreshold}{200}

\newcommand{\GateZeroItems}{43}
\newcommand{\GateZeroPct}{71.7}

\newcommand{\HaLlama}{0.276}
\newcommand{\HaMedGemma}{0.356}
\newcommand{\HaQwen}{0.122}
\newcommand{\LpHaLlama}{0.274}
\newcommand{\LpHaMedGemma}{0.360}
\newcommand{\LpHaQwen}{0.266}
\newcommand{\MarginLlama}{-4.2}
\newcommand{\MarginMedGemma}{3.8}
\newcommand{\MarginQwen}{-19.6}
\newcommand{\NBelowBaseline}{2}
\newcommand{\NEval}{500}
\newcommand{\NModels}{3}

\newcommand{\ParseEnQwen}{1.00}
\newcommand{\ParseHaLlama}{1.00}

\newcommand{\ParseHaQwen}{0.47}
\newcommand{\PoolRemaining}{1936}
\newcommand{\PoolStart}{4183}
\newcommand{\RecHiLlama}{0.63}
\newcommand{\RecHiMedGemma}{0.45}
\newcommand{\RecHiQwen}{0.77}
\newcommand{\RecLlama}{0.51}
\newcommand{\RecLoLlama}{0.38}
\newcommand{\RecLoMedGemma}{-0.01}
\newcommand{\RecLoQwen}{0.57}
\newcommand{\RecMedGemma}{0.24}

\newcommand{\RecPMedGemma}{0.073}

\newcommand{\RecQwen}{0.67}
\newcommand{\SensAN}{499}
\newcommand{\SensAShift}{0.1}
\newcommand{\SensBN}{405}
\newcommand{\SensBShift}{2.8}
\newcommand{\ValAdequacyOne}{4.52}
\newcommand{\ValAdequacyTwo}{4.36}
\newcommand{\ValFlagged}{1}

\newcommand{\ValGeFourOnePct}{100.0}

\newcommand{\ValGeFourTwoPct}{98.7}
\newcommand{\ValKappa}{-0.30}
\newcommand{\ValMinGeFourPct}{98.7}
\newcommand{\ValN}{75}

\title{Medical question answering collapses in Hausa:\\
a pre-registered pilot on three small open models}

\author{Abbas Sani\\
  Mewar University\\
  \texttt{abbassani721@gmail.com}}
\date{\today}

\begin{document}
\maketitle

\begin{abstract}
A system that reads nothing and always answers the most frequent letter scores
\BaselineP{} on our Hausa medical multiple-choice set. \NBelowBaseline{} of the
\NModels{} small open models we evaluate do not beat it. We lead with that
because the usual summary of a multilingual result -- a percentage-point drop
from English -- does not say whether what survives the drop is performance at
all.

On a paired English--Hausa set of \NEval{} MedMCQA items, three instruction-tuned
models of 3--4B parameters lose between \GapMedGemma{} and \GapQwen{} points of
accuracy, every gap significant after Holm correction. Back-translating the
Hausa items into English recovers between half and two-thirds of the gap for the
two general-purpose models but not demonstrably for the medically adapted one,
so the deficit is substantially, though not wholly, a matter of the language of
presentation. Answers are also far less stable: agreement across three option
orderings falls from \ConsEnQwen{} to \ConsHaQwen{} in the worst case. An
exploratory error analysis of \ENRows{} discordant items attributes most of the
damage to medical vocabulary lost in translation (\ETerm{}, 95\% CI
\ETermLo{}--\ETermHi{}), usually in the question stem.

A fourth registered hypothesis proposed closing the gap with self-generated
preference pairs. Its pilot gate found nothing to train on: asked to reason in
Hausa rather than to pick a letter, the model echoes the prompt back, switches
into Swahili, or simply stops without answering; only \GateParse{} of chains
are parseable. We
report that intervention as not run, which is itself the sharper result -- a
forced-choice benchmark cannot see a model that has no generative command of
the language at all.

All hypotheses, thresholds and decision rules were registered before any model
was run, and all deviations are logged.
\end{abstract}

\section{Introduction}

Large language models answer medical licensing questions well enough that the
result is no longer surprising~\cite{singhal2023clinical}. They do so in
English. Hausa is among the most widely spoken languages in Africa, 
with roughly 80 million speakers~\cite{ethnologue}, and it has almost no
medical NLP evaluation at all. The two facts are usually reported in separate
literatures, and the question that joins them is practical: if a clinician or a
patient in northern Nigeria asks one of these models a medical question in
Hausa, what is left of the ability the English benchmark measured?

We ask this of small open models specifically -- three instruction-tuned systems
of 3--4B parameters, including one medically adapted. Frontier models are
interesting, but the models that can actually be deployed on local hardware, in
a clinic without reliable connectivity or a budget for API calls, are the small
open ones. They are also the ones that AfriMed-QA found struggle most on
African medical content~\cite{olatunji2025afrimedqa}.

The study is pre-registered. Hypotheses, thresholds and decision rules were
fixed and committed before any model was run, and every departure from them is
logged in the repository. This matters here because the finding is a
deflationary one. The hypotheses themselves were supported, emphatically; what
the registered baseline reveals is that the surviving Hausa performance is
mostly not performance at all. A result of that shape is the easiest kind to
soften after the fact, by choosing a comparison that flatters it.

Our contributions are: (i) a paired English--Hausa medical MCQ set of \NEval{}
items with segment-level quality estimation and two-rater human validation,
released with the annotation sheets; (ii) a pre-registered evaluation of three
small open models with decision rules fixed in advance, reporting accuracy,
answer stability under option reordering, and a translate-test probe; and
(iii) the observation that the conventional way of reporting a multilingual gap
-- as a percentage drop -- conceals whether the post-drop score is above a
trivial baseline. Here, for \NBelowBaseline{} of the \NModels{}, it is not.

\section{Related work}

\paragraph{Medical question answering, in English.}
MedMCQA~\cite{pal2022medmcqa} is one of the standard multiple-choice medical QA
sets, drawn from Indian postgraduate entrance examinations, and is the source we
sample from. MultiMedQA assembled six such datasets and showed that a large
instruction-tuned model reaches a passing standard on USMLE-style
questions~\cite{singhal2023clinical}. Both are English throughout, as is almost
all of the medical QA literature these results come from. What they establish is
a ceiling: whatever a model can do medically, it can do it in English.

\paragraph{African languages, and African medical content.}
IrokoBench~\cite{adelani2025irokobench} human-translated reasoning and knowledge
benchmarks into seventeen African languages and found large gaps between English
and African-language performance across ten open and four proprietary models,
with the strongest open model reaching a fraction of the best proprietary one.
AfriMed-QA~\cite{olatunji2025afrimedqa} approaches the problem from the other
side: it collects medical questions authored across more than sixty African
medical schools, capturing clinical content and disease prevalence that
Western-sourced benchmarks miss, and reports that smaller edge-friendly models
struggle to reach a passing score.

These two leave a gap that this study occupies. IrokoBench covers African
languages but general knowledge and reasoning rather than medicine; AfriMed-QA
covers African medical content but poses it in English. Neither asks what
happens to medical question answering when the question itself is in an African
language. That is the question here, for one language, in depth, with the
hypotheses fixed in advance.

\paragraph{Translate-test.}
Translating a non-English input into English and reasoning there is a standard
strong baseline in multilingual evaluation~\cite{shi2023multilingual}, and
IrokoBench finds it helps larger English-centric
models~\cite{adelani2025irokobench}. We use it differently. H2 back-translates
the Hausa items into English and re-asks, not to maximise a score but as a
diagnostic: the fraction of the gap it recovers separates the part of the
deficit that travels with the language of presentation from the part that does
not. A model that recovers fully was never short of the medical knowledge; one
that does not recover has lost something the round trip cannot restore.

\paragraph{Multiple-choice evaluation is fragile.}
Reported MCQ accuracy is less stable than it looks. Zheng et
al.~\cite{zheng2024selection} show that models carry a selection bias toward
particular option identifiers -- largely a token bias on the letters themselves
rather than a positional one -- which makes a score depend on where the correct
answer sits. Pezeshkpour and Hruschka~\cite{pezeshkpour2024order} report
substantial swings on several benchmarks from reordering the options alone, and
argue the effect concentrates where the model is uncertain between its top
candidates.

Three of our design choices follow directly. We report what a model scores by
answering the most frequent letter and reading nothing, because a letter bias
and a non-uniform answer key together make that a non-trivial score. We present
every item in three option orders and register agreement across them as an
outcome in its own right, since a model that is merely guessing will answer
inconsistently even when its accuracy looks respectable. And we report a
secondary scoring rule based on the probability of each letter token alongside
the primary parsed one, which separates not knowing the answer from not
following the answer format.

\paragraph{Preference optimisation.}
H4 asks whether self-generated preference pairs can close part of the gap, using
direct preference optimisation~\cite{rafailov2023dpo} with hyperparameters
reused unchanged from the authors' prior study, so the two remain comparable.

\section{Data}

We sample from the MedMCQA validation split, whose labels are public. Of
\PoolStart{} items we remove Dental (\ExclDental{} items, outside this study's
general-health motivation and about 30\% of the split), multi-answer items
(\ExclMultiAnswer{}), items whose options are position-dependent (``all of the
above'' and similar), and items that require an image the model cannot see,
leaving \PoolRemaining{}. From these we draw \NEval{} evaluation items stratified
by subject with a fixed seed. MedMCQA contains a scraping artifact in which the
letter pair ``rt'' is deleted from many words (``aery'' for artery); we repair it
with a fixed rule list before filtering.

Hausa versions are produced with NLLB-200-3.3B~\cite{nllb2022}. Segment-level quality is
estimated by round-trip chrF1~\cite{popovic2015chrf} with a second, distilled model, and flagged
segments are reviewed by a native speaker who records one of \emph{keep},
\emph{english} (use the English term, as is common for medical vocabulary in
Hausa) or \emph{edit}. \FallbackItems{} items retain at least one English option
after review; a pre-registered sensitivity analysis excludes them.

\paragraph{Human validation.}
Two native speakers independently rated \ValN{} items for adequacy on a 1--5
scale. Mean adequacy was \ValAdequacyOne{} and \ValAdequacyTwo{}, with
\ValGeFourOnePct{}\% and \ValGeFourTwoPct{}\% of items rated 4 or better, and no
medical-term errors marked by either rater. Quadratic-weighted $\kappa$ is
\ValKappa{}: ratings concentrate at 4--5, where $\kappa$ is unstable, so we
report it as computed and rely on the agreement that matters -- both raters
judged at least \ValMinGeFourPct{}\% of the items acceptable. \ValFlagged{} item
was flagged by either rater.
Both raters reported checking question stems closely and options less closely,
so this validates stem quality better than option quality; we return to this in
the limitations.

\paragraph{Licensing.}
MedMCQA carries two different licences from its own publishers: the source
project is MIT, while the Hugging Face distribution's metadata says
\texttt{apache-2.0} and its licensing section is an unfilled placeholder. Both
permit redistribution of derived text with attribution. The binding constraint
is neither: NLLB-200-3.3B is CC-BY-NC-4.0, so the released Hausa text is
non-commercial only.

\section{Method}

\paragraph{Pre-registration.}
Hypotheses H1--H4, their thresholds and their decision rules were frozen before
any model was run, and are in the repository along with every deviation. Six
deviations are logged; the two that affect the numbers below are an
assistant-turn prefill (the answer turn begins with ``Answer:'' / ``Amsa:'', so
the model's first tokens are the letter) and a secondary letter-probability
scoring rule reported alongside the primary parsed one.

\paragraph{Models and conditions.}
Llama-3.2-3B-Instruct~\cite{llama32}, Qwen2.5-3B-Instruct~\cite{qwen25} and
MedGemma-4B-IT~\cite{medgemma}, all bf16. Each item is presented in English and in Hausa
under a fixed zero-shot template, in the canonical option order and in two seeded reorderings,
plus a back-translated-to-English condition for H2.

\paragraph{Statistics.}
Paired comparisons use an exact McNemar test~\cite{mcnemar1947} with Holm
correction~\cite{holm1979} across the three models, and 95\% confidence intervals from a paired bootstrap with 10,000
resamples. Consistency (H3) is the proportion of items answered identically
across all three option orders.

\paragraph{The baseline that matters.}
Gold answers are not uniformly distributed over A--D, so we report what a model
scores by reading nothing and always answering the most frequent letter:
\BaselineP{} in the canonical order, against \Chance{} for chance. Every accuracy
below should be read against it.

\section{Results}

\begin{figure}[t]
  \centering
  \includegraphics[width=\linewidth]{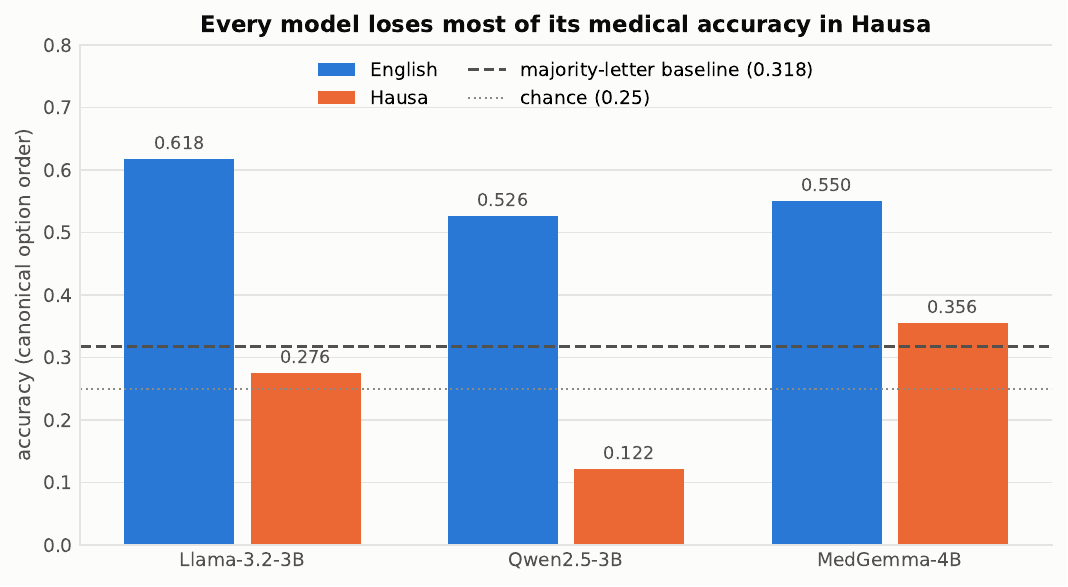}
  \caption{English and Hausa accuracy for each model, against the
  majority-letter baseline (\BaselineP{}) and chance. Two of the three models
  fall at or below the baseline in Hausa.}
  \label{fig:accuracy}
\end{figure}

\paragraph{H1: the gap.}
All three models lose a large fraction of their accuracy
(Figure~\ref{fig:accuracy}). Llama falls from \EnLlama{} to \HaLlama{}
(\GapLlama{} pp, 95\% CI \GapLoLlama{}--\GapHiLlama{}), Qwen from \EnQwen{} to
\HaQwen{} (\GapQwen{} pp, \GapLoQwen{}--\GapHiQwen{}), and MedGemma from
\EnMedGemma{} to \HaMedGemma{} (\GapMedGemma{} pp,
\GapLoMedGemma{}--\GapHiMedGemma{}). All three are significant after Holm
correction.

\paragraph{The gap is not the headline.}
\NBelowBaseline{} of the \NModels{} models are at or below the \BaselineP{}
majority-letter baseline in Hausa: Llama at \HaLlama{} (\MarginLlama{} points)
and Qwen at \HaQwen{} (\MarginQwen{}). Only MedGemma (\HaMedGemma{}) is above
it, by \MarginMedGemma{} points. Reporting these systems as retaining ``some''
medical ability in Hausa would be an artifact of comparing against chance rather
than against the strongest trivial strategy available. The comparison is not a
rhetorical device: the majority-letter strategy is what the selection bias
documented by Zheng et al.~\cite{zheng2024selection} converges on when a model
has nothing else to go on.

\begin{figure}[t]
  \centering
  \includegraphics[width=\linewidth]{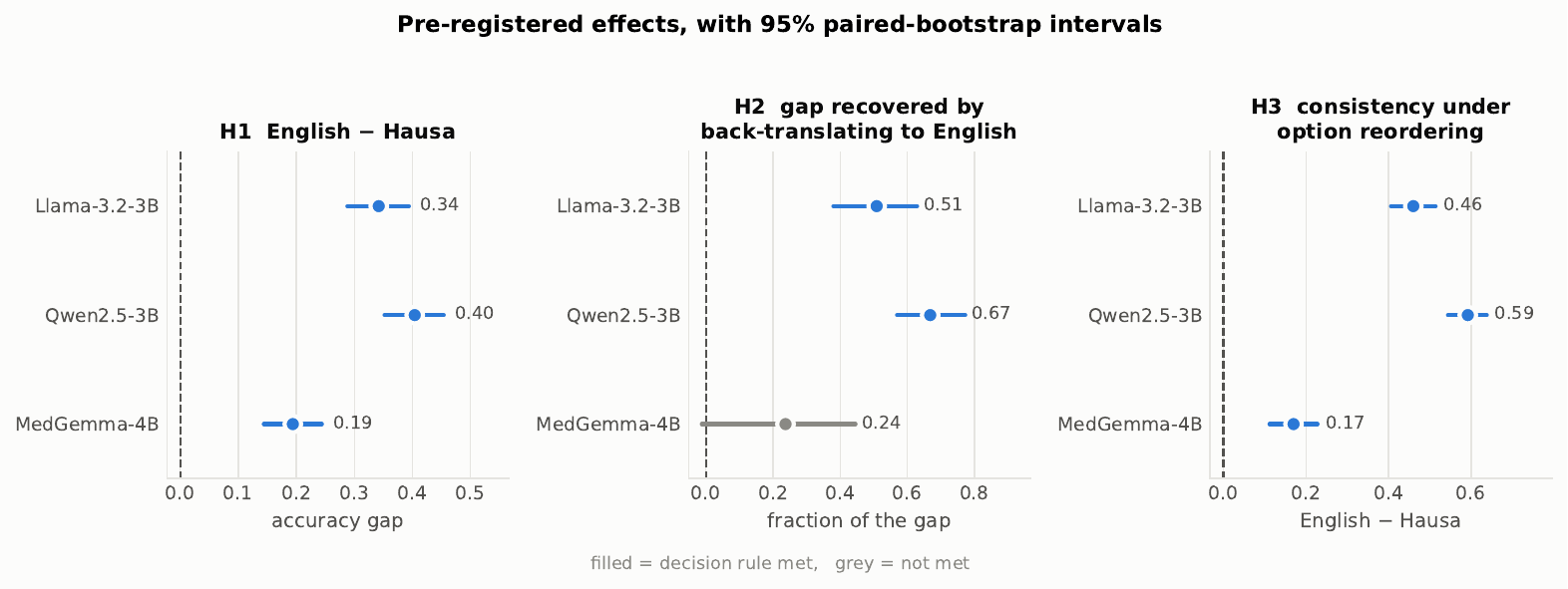}
  \caption{The three pre-registered effects with 95\% paired-bootstrap
  intervals. Grey marks the one decision rule of nine that is not met.}
  \label{fig:effects}
\end{figure}

\paragraph{H2: back-translation.}
Translating the Hausa items back into English and re-asking recovers
\RecLlama{} of Llama's gap (95\% CI \RecLoLlama{}--\RecHiLlama{}) and \RecQwen{}
of Qwen's (\RecLoQwen{}--\RecHiQwen{}), but only \RecMedGemma{} of MedGemma's,
with an interval spanning zero (\RecLoMedGemma{}--\RecHiMedGemma{},
$p=\RecPMedGemma{}$). H2 is therefore supported for two models and not for the
third. Where it holds, much of the deficit travels with the language of
presentation rather than with the medical content.

\paragraph{H3: stability.}
Agreement across the three option orders falls from \ConsEnLlama{} to
\ConsHaLlama{} for Llama, \ConsEnQwen{} to \ConsHaQwen{} for Qwen, and
\ConsEnMedGemma{} to \ConsHaMedGemma{} for MedGemma. Supported for all three.
A Hausa answer is not only likelier to be wrong; it is likelier to change when
the options are shuffled.

\begin{figure}[t]
  \centering
  \includegraphics[width=\linewidth]{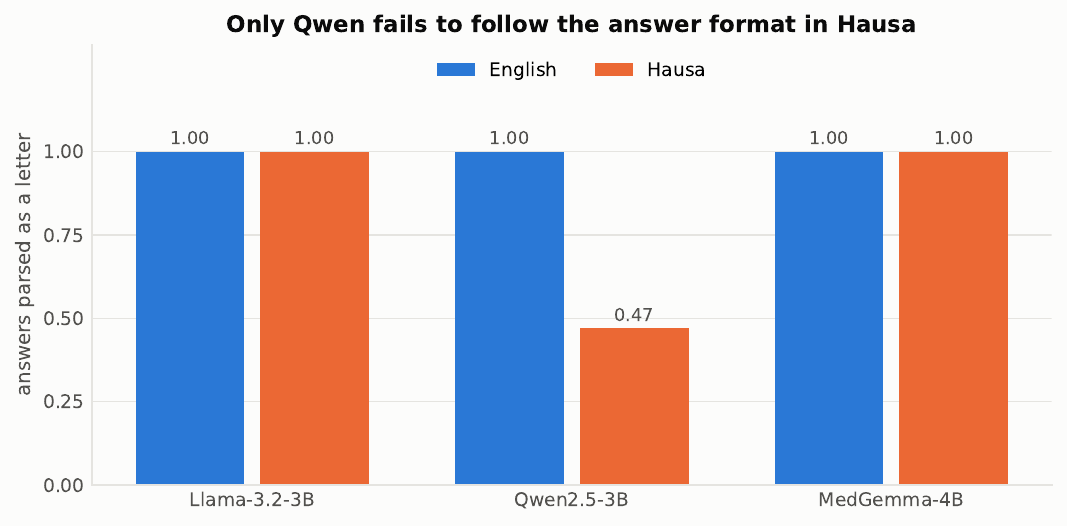}
  \caption{Proportion of answers that could be parsed as a letter. Qwen is the
  exception, and its Hausa deficit is partly a format-following failure.}
  \label{fig:parse}
\end{figure}

\begin{figure}[t]
  \centering
  \includegraphics[width=\linewidth]{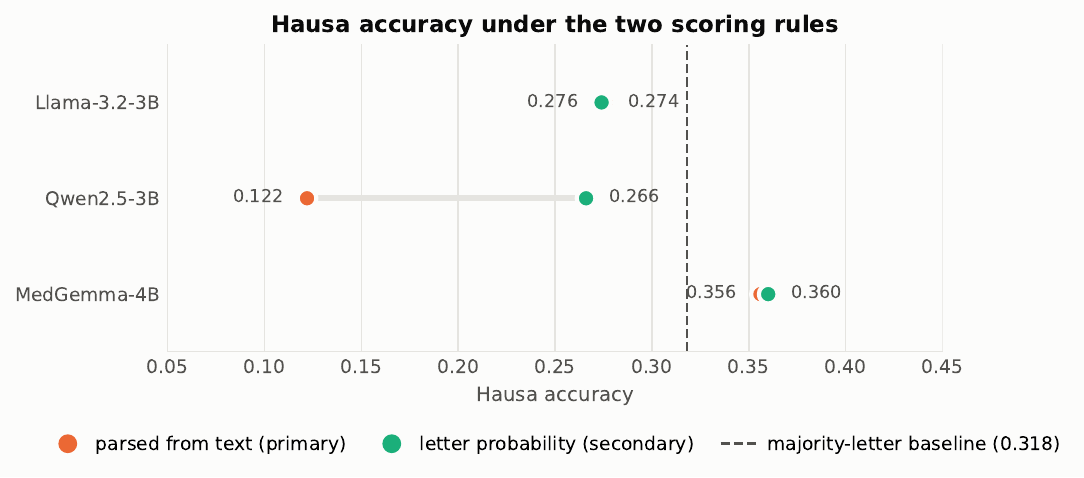}
  \caption{Hausa accuracy for the same model outputs under the primary
  (text-parsed) and secondary (letter-probability) scoring rules. Only Qwen
  moves appreciably.}
  \label{fig:scoring}
\end{figure}

\paragraph{Format following, and what the scoring rule carries.}
Llama and MedGemma emit a parseable letter for every Hausa item. Qwen parses
only \ParseHaQwen{} of its Hausa answers against \ParseEnQwen{} in English
(Figure~\ref{fig:parse}). Scoring the identical outputs by the probability of
each letter token rather than by parsing the text moves Qwen from \HaQwen{} to
\LpHaQwen{}, while Llama (\HaLlama{} to \LpHaLlama{}) and MedGemma
(\HaMedGemma{} to \LpHaMedGemma{}) barely move (Figure~\ref{fig:scoring}).
Reporting only the parsed rule overstates Qwen's deficit; reporting only letter
probability hides a real failure to follow instructions in Hausa. We report
both. Note that even under the more generous rule Qwen remains below the
\BaselineP{} baseline.

\paragraph{Sensitivity.}
Both pre-registered sensitivity analyses leave H1 intact. Dropping items flagged
by either rater (\SensAN{} remain) moves no model's gap by more than
\SensAShift{} points, and dropping items that retain an English option
(\SensBN{} remain) by no more than \SensBShift{}. The gap is therefore not an
artifact of the handful of items where the translation was known to be weak.

\begin{figure}[t]
  \centering
  \includegraphics[width=\linewidth]{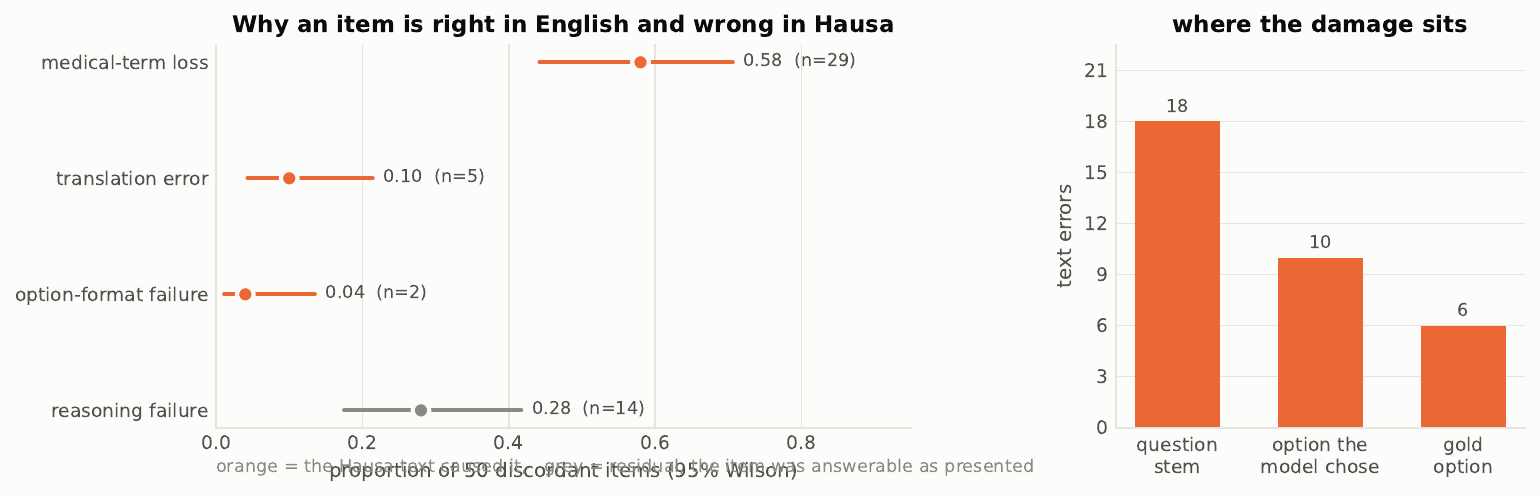}
  \caption{Exploratory error analysis of \ENRows{} items answered correctly in
  English and wrongly in Hausa.}
  \label{fig:errors}
\end{figure}

\paragraph{Exploratory: why items fail.}
We coded \ENRows{} discordant items -- correct in English, wrong in Hausa --
into four categories fixed in advance, under a written codebook with two
machine-enforced consistency rules: a coded segment's Hausa must differ from its
English, and a term or translation error must sit in the question stem, the gold
option or the option the model chose. Medical-term loss accounts for \ETerm{}
(95\% CI \ETermLo{}--\ETermHi{}, $n=\ETermN{}$), residual reasoning failure for
\EReason{}, translation error for \ETrans{} and option-format failure for
\EFormat{}. Of the \ETextTotal{} text errors, \ELocStem{} sit on the question
stem, \ELocChosen{} on the option the model chose and \ELocGold{} on the gold
option. Separately, \EUntranslated{} of the \ENRows{} items carry at least one
option whose Hausa is identical to its English -- untranslated medical
vocabulary, recorded as a descriptive covariate rather than as damage, since the
model sees the same string either way.

This analysis is exploratory, with no hypothesis attached. One coder coded all
items and a second native speaker verified that coding; it is not independent
double coding, and we therefore report no $\kappa$.

\section{H4: can self-generated preference pairs close the gap?}

H4 asked whether preference pairs generated by the model itself could close
part of the Hausa gap. The protocol samples \GateK{} chain-of-thought
completions per training item, keeps items where some chains reach the gold
answer and others do not, and trains on the resulting (chosen, rejected) pairs.
Before committing to it we ran a pre-registered pilot on \GateItems{} items
(\GateSamples{} chains) to check that the procedure had anything to work with.
It did not, and we report H4 as not run.

\paragraph{The gate.}
Only \GateParse{} of chains stated an answer letter at all. Of the
\GateItems{} items, \GateZeroItems{} (\GateZeroPct{}\%) had no correct chain
among their \GateK{}, and none had more than two. Pair yield was
\GatePairs{} items, projecting to \GateProjected{} over the full pool against a
pre-registered floor of \GateThreshold{}. Accuracy over all chains was
\GateAcc{}; restricted to the \GateParsedN{} chains that do state a letter it is
\GateAccParsed{} (95\% Wilson \GateAccParsedLo{}--\GateAccParsedHi{}), an
interval that contains \Chance{}. The model is therefore indistinguishable from
chance on the chains it manages to complete, and silent on the rest. Under the
estimator $\mathrm{acc} = \mathrm{skill} + (1-\mathrm{skill})/4$ the implied
skill rate is zero, meaning the ``chosen'' side of every pair is consistent with
a lucky guess. Training on such pairs would teach a preference for guessing.

\paragraph{Why the chains fail.}
We classified all \FailChains{} chains automatically
(Table~\ref{tab:failmodes}). Two modes dominate and a third, which an informal
reading of a handful of chains had suggested, turns out to be negligible.

\begin{table}[t]
  \centering
  \begin{tabular}{lrr}
    \toprule
     & all \FailChains{} chains & the \FailUnparsed{} unparseable \\
    \midrule
    echoes the prompt        & \FailAllPromptEchoN{} (\FailAllPromptEcho{}\%)
                             & \FailAnyPromptEchoN{} (\FailAnyPromptEcho{}\%) \\
    carries a Swahili marker & \FailAllSwahiliN{} (\FailAllSwahili{}\%)
                             & \FailAnySwahiliN{} (\FailAnySwahili{}\%) \\
    cut off at the budget    & \FailAllTruncatedN{} (\FailAllTruncated{}\%)
                             & \FailAnyTruncatedN{} (\FailAnyTruncated{}\%) \\
    degenerate repetition    & \FailAllRepetitionN{} (\FailAllRepetition{}\%)
                             & \FailAnyRepetitionN{} (\FailAnyRepetition{}\%) \\
    \bottomrule
  \end{tabular}
  \caption{Failure modes among the pilot's chains. Categories overlap. Swahili
  detection uses a marker list that excludes words Hausa and Swahili share, so
  it is a lower bound.}
  \label{tab:failmodes}
\end{table}

First, \FailAllPromptEcho{}\% of chains \textbf{echo the prompt} rather than
answering it -- restating the instruction, or emitting a fresh question stem
with its own option list. Second, \FailAllSwahili{}\% \textbf{code-switch out of
Hausa}, almost always into Swahili: chains close ``Jibu la sahihi ni C'' or
``Jibu: X = A'' instead of the requested \emph{Amsa}, and carry Swahili
function words (\emph{hivyo}, \emph{ambayo}, \emph{kwa hivyo}) through otherwise
Hausa text. Swahili is the African language best represented in the model's
training data, and it appears to be what the model falls back to when asked to
produce Hausa. Because the marker list excludes vocabulary the two languages
share, this is a lower bound.

Neither mode is an artifact of the generation budget. The median chain is
\FailMedianTokens{} tokens against a cap of \FailTokenBudget{}, only
\FailAtBudget{} of \FailChains{} chains reach it, and exactly
\FailAllRepetitionN{} chain degenerates into repetition. The model is not
running out of room to reason; it stops early, having produced something other
than Hausa reasoning.

The same pathologies appear in the chains that \emph{do} yield a letter:
\FailParsedPromptEcho{}\% of them echo the prompt and
\FailParsedCodeSwitch{}\% carry Swahili. A parseable answer here is typically a
bare assertion rather than a derivation, which is why the pairs carry no signal
even where they exist.

This is the substantive finding, and it is not the one H4 was designed to
produce. Our baseline evaluation constrains the model to emit a single letter
after a prefill, and under that constraint Llama parses \ParseHaLlama{} of its
Hausa answers. Asked instead to generate several sentences of Hausa, the same
model cannot sustain the language. The deficit measured in
Section~\ref{sec:results} is thus a lower bound on the practical one: a system
that must \emph{explain} in Hausa fails in ways that a forced-choice benchmark
cannot see.

\paragraph{What we did not do.}
We did not substitute a different intervention model, retune the prompt until
the gate passed, or lower the pair threshold. H1--H3 were already known when
the gate ran, and any of those moves would be selecting a method by its
outcome. Raising $k$ from \GateK{} to 8 is pre-registered as the one permitted
response to a low pair count, but it addresses yield, not the zero skill rate
that makes the yield worthless, so we did not spend the compute. The gate
report is released as \texttt{results/h4\_pilot\_gate.json} and the inspected
chains as \texttt{results/h4\_pilot\_chains.csv}.

\paragraph{A note on the ceiling.}
Even a fully successful H4 would not have changed this paper's conclusion. The
pre-registered minimum gain is 3 points on a base of \HaLlama{}, which reaches
0.306 -- still below the \BaselineP{} majority-letter baseline.

\section{Limitations}

\paragraph{The Hausa set is machine-translated.}
This is the central caveat and it cuts in a specific direction. Our own error
analysis attributes most discordance to the translation rather than to the
model. The measured gap is therefore best read as \emph{the gap a Hausa speaker
would encounter using these models through the best freely available
translation} -- which is the practically relevant quantity -- and not as a clean
measure of multilingual medical reasoning. The human validation subset bounds how
much is translation noise, and both sensitivity analyses leave H1 unchanged.

\paragraph{Option quality is less well validated than stem quality.}
Both raters reported checking stems closely and options less closely.

\paragraph{Scale.}
\NEval{} items, three models of 3--4B parameters, one language. This is a pilot
and is reported as one.

\paragraph{The error analysis is single-coder.}
With a verification pass, under a written codebook with enforced rules, but not
independently double-coded.

\section{Conclusion}

Posed in Hausa, three small open models -- two general-purpose, one medically
adapted -- lose most of their accuracy and much of their stability, and
\NBelowBaseline{} of the \NModels{} end up at or below a baseline that reads
nothing at all. Where the Hausa text is the cause, it is usually medical
vocabulary, and usually in the question stem. For the two general-purpose
models, back-translating into English recovers much of the gap, which points at
the language of presentation rather than the medicine as the bottleneck; for the
medically adapted one it does not, and we cannot say from these data whether
that is because its deficit is of a different kind or because its smaller gap
leaves less to recover.

The intervention we had registered to close the gap never ran, because its
pilot found no signal to train on. That failure is the most useful thing in the
study. Constrained to emit a letter, these models look merely weak in Hausa;
asked to produce Hausa, one of them cannot. Benchmarks built on forced choice
will keep reporting the first number and missing the second. We release the
paired set, the human annotation sheets, the pre-registration and every
deviation.

\section*{Acknowledgements}
We thank Mr Rufa'i Yusuf Zakari for the second translation-adequacy rating and
Usman Haruna for the verification pass on the error coding, both of whom consented to being named.

\section*{Data and code availability}
Code, pre-registration, deviation log, human annotation sheets and all result
files: \url{https://github.com/avvas200/hausa-med-qa}. The derived Hausa text is
CC-BY-NC-4.0, inherited from NLLB-200.

\bibliographystyle{plain}
\bibliography{refs}

\end{document}